# CNN-MERP: An FPGA-Based Memory-Efficient Reconfigurable Processor for Forward and Backward Propagation of Convolutional Neural Networks

Xushen Han, Dajiang Zhou, Shihao Wang, and Shinji Kimura
Graduate School of Information, Production and Systems, Waseda University
2-7 Hibikino, Wakamatsu-ku, Kitakyushu, Fukuoka 808-0135, Japan
hanxushen@fuji.waseda.jp

*Abstract*—Large-scale deep convolutional neural networks (CNNs) are widely used in machine learning applications. While CNNs involve huge complexity, VLSI (ASIC and FPGA) chips that deliver high-density integration of computational resources are regarded as a promising platform for CNN's implementation. At massive parallelism of computational units, however, the external memory bandwidth, which is constrained by the pin count of the VLSI chip, becomes the system bottleneck. Moreover, VLSI solutions are usually regarded as a lack of the flexibility to be reconfigured for the various parameters of CNNs. This paper presents CNN-MERP to address these issues. CNN-MERP incorporates an efficient memory hierarchy that significantly reduces the bandwidth requirements from multiple optimizations including on/off-chip data allocation, data flow optimization and data reuse. The proposed 2-level reconfigurability is utilized to enable fast and efficient reconfiguration, which is based on the control logic and the multiboot feature of FPGA. As a result, an external memory bandwidth requirement of 1.94MB/GFlop is achieved, which is 55% lower than prior arts. Under limited DRAM bandwidth, a system throughput of 1244GFlop/s is achieved at the Vertex UltraScale platform, which is 5.48 times higher than the state-of-the-art FPGA implementations.

## I. INTRODUCTION

State-of-the-art CNNs are demonstrating high performance and flexibility in a wide range of applications including video surveillance, object recognition and mobile robot vision [1]–[3]. In recent years there is a growing tendency to take advantage of large-scale and deep convolutional neural networks. For example, in AlexNet [4], a deep CNN plays a crucial role in significantly improving image identification accuracy. A CNN is generally composed of several cascades of convolutional, activation and pooling layers, with an example given in Figure 1. Currently, the depth and the number of feature maps in layers keep increasing, which lead to tremendous computing workloads, in both the forward and backward propagation phases. With two NVIDIA GTX 580 3GB GPUs, the training of AlexNet still took about six days to finish the 90 epochs, which are necessary for a satisfactory model.

Many of today's deep CNNs are implemented on GPU platforms, based on which some fast and friendly frameworks are developed such as Caffe [5], Torch [6], and Chainer [7]. These frameworks are designed for easy adjustment of the structures of neural networks. From the implementation's point of view however, dedicated architectures for CNNs have a better potential for a higher throughput at the same area as well as a higher energy efficiency. Hence FPGA and ASIC implementations have come to researchers' attention. Previous FPGA/ASIC architectures already achieved a throughput of several hundreds of Gop/s. The computational components of these architectures are also extensible, i.e. higher performance can be achieved by leveraging parallelism. The current Xilinx VU13P FPGA already contains over 10 thousand DSPs, which are promising for delivering a 4TFlop/s throughput at a moderate 200MHz clock rate. However, most of these designs are still faced with two problems:

1) While the density of VLSI chips allows a massive parallelism of computation, external memory bandwidth becomes the system bottleneck.

2) With deep CNNs comprising many layers with different characteristics, it is particularly difficult to make a single architecture reusable by and optimum for all the layers.

In this paper, we present CNN-MERP, an FPGA-based CNN processor for both forward and backward propagation, addressing the above issues with the following contributions.

1) A highly memory-efficient architecture for the high-throughput CNN processor.

2) 2-level reconfigurability for forward and backward propagation of different layers in CNNs.

The rest of this paper is organized as follows. Section 2 introduces related work. We give an overview of CNNs in Section 3. Strategies of external bandwidth reduction are shown in section 4. The detailed hardware design and reconfigurability are introduced in Section 5 and Section 6, respectively. The results of implementation are given in Section 7 and the final section is for conclusion and future work.

## II. RELATED WORK

FPGA-based CNN acceleration has been discussed in several previous papers. Chakradhar, et al. [8]'s 16-bit processor can automatically analyze workloads and configure hardware for higher speed. To optimize the throughput with limited memory bandwidth, a roofline-model-based design was proposed by Chen Zhang, et al [9]. Their implementation can reach 61.62Gflop/s with a 1.55GB/s external memory. To optimize the usage of on-chip memory, in 2013, Maurice et al [18] take advantage of high-level synthesis tools to develop a configurable accelerator template, which comprises 140MACC with 150MB/s memory. In 2011, NeuFlow, an 160Gop/s processor was published by Farabet Clément, et al. [10]. Another implementation is nn-X, proposed by Vinayak Gokhale, et al. [11]. This processor achieves 227Gop/s with the 7.6GB/s DMA channels.

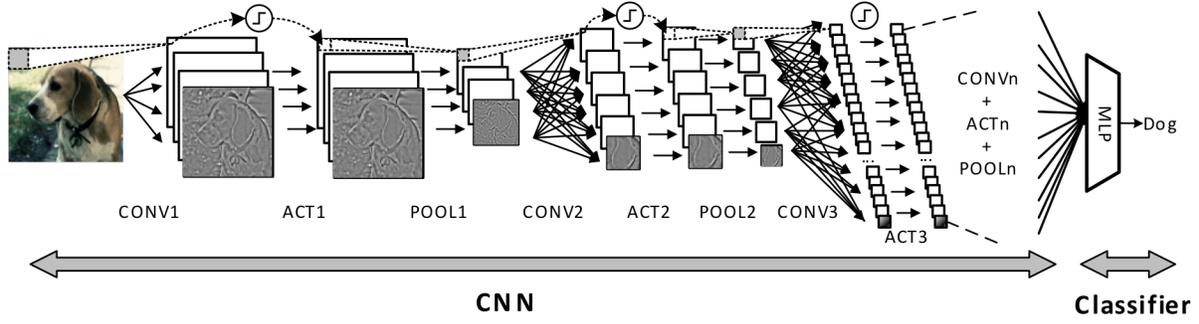

Figure 1 Example of convolutional neural network. CONV: convolutional layer. ACT: activation layers. POOL: pooling layer. For AlexNet, the number of input and output feature maps in layers are: CONV1+ACT1+POOL1: 3→96, CONV2+ACT2+POOL2: 48→128, CONV3+ACT3: 256→384, CONV4+ACT4→192→192, CONV5+ACT5+POOL5: 192→128.

Another choice for accelerating CNNs is to use ASIC implementation, e.g. NeuFlow by Phi-Hung Pham, et al [12], Diannao by Tianshi Chen, et al [13], Origami by Lukas Cavigelli [14], and [15]. Since ASICs provide higher density, the bottleneck of the external memory is even more critical. To reduce the traffic, one idea is storing all feature maps and kernels on chip. In Dadiannao, Chen et al. [16] proposed to store the whole network on chip with eDRAM. Such a design style, however, imposes a critical constraint on the scale of CNNs that can be implemented, which makes it less practical for ultra-large-scale networks. Another idea is reusing input data. In 2016, Yu-Hsin Chen, et al [17] published a processor Eyeriss, which takes advantage of an 108KB on-chip memory to reduce normalized bandwidth requirement to 4.31MB/Gop.

If we apply the current design style into a training process of a large-scale CNNs, e.g. the AlexNet [4], 32-bit operands and more processing elements are necessary, since the requirement for higher precision and workloads have been demonstrated in training. While a high-performance FPGA like VU13P enables a maximum computational throughput of 4TFlop/s, the requirement of memory bandwidth correspondingly reaches *4.31MB/Gop×4Top/s×(32b/16b)= 34.48GB/s*, which is too heavy to attain with current mainstream DDR4 (19.2GB/s) memory.

## III. CONVOLUTIONAL NEURAL NETWORKS

Typically, modern CNNs are composed of some basic layers, i.e. convolutional layers, activation layers and pooling layers. The Training process updates kernels to build up specific functionalities via 2 phases: the forward propagation (FP), and the backward propagation (BP). The FP focuses on the prediction of input images, while the BP updates kernels with corresponding partial derivatives of a cost function.

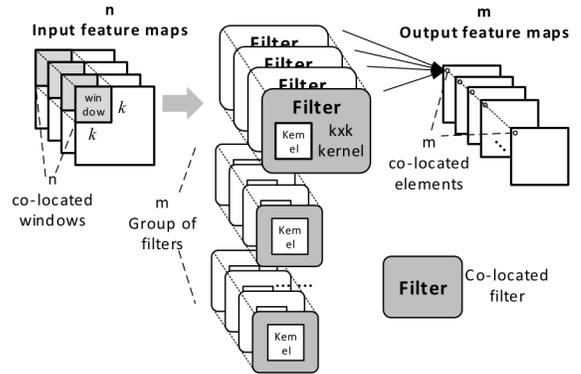

Figure 2 convolutional layer in forward propagation phase. Grey filters are composed of co-located filters. n co-located windows and n filters in one group determine one element in output feature map. m group of filters corresponds m co-located output elements

### A. Forward propagation phase

Convolutional layers accept n input feature maps to produce m output feature maps. As shown in Figure 2, $k \times k$ elements in an input feature map are combined into an input window, which is passed through a filter. A filter is defined as the convolutional operations for a pair of input window and kernel. The output of a filter is a partial result of output element, which is shown in following equation.

$$y_{partial}[r,c] = \sum_{i=0}^{k-1}\sum_{j=0}^{k-1} x[r+i, c+j] \cdot Ker[i,j],$$

where [r,c] corresponds coordinates in the output feature map $y$. Ker means kernels. A Group of filters contributes the same output element of $y$, while Co-located filters are defined as the filters in different groups concerning the same window.

Activation layers provide non-linearity using activation functions. Currently, a popular choice of activation function is the Rectified Linear Units (ReLU). In [19], Vinod Nair, et al stated that ReLU has better generalized ability to prevent training model from saturating. In the meanwhile, the ReLU is easy to implement compared to other exponent-based functions.

Pooling layers compute the average value of $p \times p$ neighboring elements in the same feature map. As shown in Figure 1, there is no interconnection among different feature maps so they can compute subsampling results independently.

## B. Backward propagation phase

The BP phase includes two steps: $\delta$ propagation and kernel updating. $\delta$ is defined as the partial derivative of the cost function $J$ with respected to the output feature map $y$ in FP. $\delta'$s can be determined via propagations due to the propagating relation of $y$ between layers and the chain rule of the partial derivatives. Since each element of $y$ owns derivative, the dimensions of $\delta$'s and $y$'s keep the same. However, the calculations of $\delta$ propagation are different.

Convolutional layers are similar as those in FP phase. Differences include 180-degrees-rotated kernels and input $\delta$ maps with zero padding of $k$-$1$. Activation layers accept outputs of corresponding activation layers in FP to calculate derivatives of ReLU. Then the derivatives and inputs are multiplied to get outputs. Pooling layers up-sample the input $\delta$ maps. One element in inputs should be multiplied by $1/(p\times p)$, then the one result is copied to $p\times p$ neighboring outputs identically.

The kernel updating (KU) step determines the partial derivatives of kernels which are utilized for optimization. The algorithm of optimization is the gradient decent, which is shown in the following equation,

$$Ker[i,j] = Ker[i,j] - \alpha \frac{\partial J}{\partial Ker[i,j]} = Ker[i,j] - \alpha \sum_{(r,c)\in\delta} \delta[r,c] \cdot x[r+i,c+j]$$

where $\alpha$ is a constant of learning rate, $x[r+i,c+j]$ corresponds an input window. Hence the partial derivatives of kernels can be determined by windows of input feature maps and corresponding elements of $\delta$ maps.

## IV. ANALYSIS OF MEMORY TRAFFIC

### A. Memory traffic evaluation

The parallelism of computational units (CUs) determines both the throughput and the bandwidth requirement of a processor. According to the structure of the layer, filters process data independently. As a result, The CU, which is designed for operations in a filter can be arranged in parallel for higher throughput. Figure 3 shows the arrangement of three parallel CUs with a detailed design of multiplications and additions. Considering operations in a filter, all multiplications are in parallel and pipelined with additions. This parallelism leverages high throughput

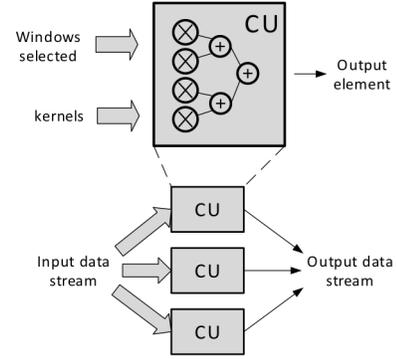

Figure 3 Parallelism within computational units (CU) and parallelism between CUs.

with the cost of heavy bandwidth requirement, which is discussed in the following section.

We use a typical convolutional layer, followed with activation and pooling layers as an example to evaluate the external memory traffic. These layers are the 2$^{nd}$ layers from the AlexNet [4] that is shown in Table 1. Two characters should be noticed. One is the convolutional layer involves the majority of computation: Three layers include 28.70G operations in total, while the convolutional layer takes up 28.67G of it. The other character is the duplicated data loading: 65.7MB off-chip data result in 117.0GB external traffic. We try to reduce the external traffic from multiple aspects including on/off-chip allocation, data flow optimization and data reuse, which are realized via the following five strategies.

### B. Strategy 1: on/off-chip data allocation

A perceived wisdom for reduction of memory traffic is to allocate a moderate amount of data on chip. Theoretically speaking, if there exists a powerful implementation which has ability to cover all data on chip, except for the data initialization, there is no external traffic required. Nevertheless, for our FPGA, such allocation is impractical because of space limitation. In the typical layer shown in Table 1, kernels require 614.4KB memory, while the demand of feature maps is *17.9MB+ 47.8MB×4+11.0MB =220.1MB*. In the meanwhile, caching all kernels and caching all feature maps contribute to the same degree of traffic reduction -- either of them reduces half of the inputs, since every multiplication only need one off-chip operand. In comparison, storing all kernels on chip is more appropriate. As a result, the external traffic for computation only comes from the feature maps.

Such a strategy can reduce half of external traffic of inputs, which makes the normalized traffic decrease to *(114.7GB/2+ 2.3GB)/28.6GFlop=2085MB/GFlop*. In turn, 614.4KB on-chip memory is needed for storage of kernels.

### C. Strategy 2: reuse input windows between filters

One window (size is $k\times k$) of input feature maps is shared by co-located filters base on the structures of CNNs. All co-located filters acquire the same window to produce intermediate results of co-located output elements. The conventional method [11] reloads one window for different filters, which aggravates external bandwidth requirement.

Table 1 Memory traffic for 2nd layers in AlexNet

| Layers | | 2$^{nd}$ conv | 2$^{nd}$ act | 2$^{nd}$ pool |
|---|---|---|---|---|
| input feature maps | Number | 48 | 128 | 128 |
| | Size | 27×27 | 27×27 | 27×27 |
| | Storage | 17.9MB | 47.8MB | 47.8MB |
| | Traffic | 114.7GB | 47.8MB | 99.7MB |
| output feature maps | Number | 128 | 128 | 128 |
| | Size | 27×27 | 27×27 | 13×13 |
| | Storage | 47.8MB | 47.8MB | 11.0MB |
| | Traffic | 2.3GB | 47.8MB | 11.0MB |
| kernels | Size | 5×5 | --- | 3×3 |
| | Storage | 614.4KB | --- | --- |
| Total | Traffic | 117.0GB | 95.6MB | 110.7MB |
| | operations | 28.67G | 11.9M | 24.9M |

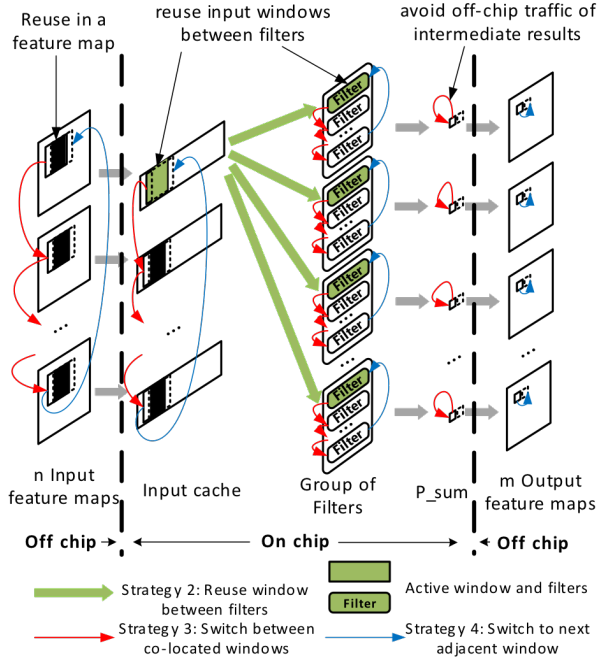

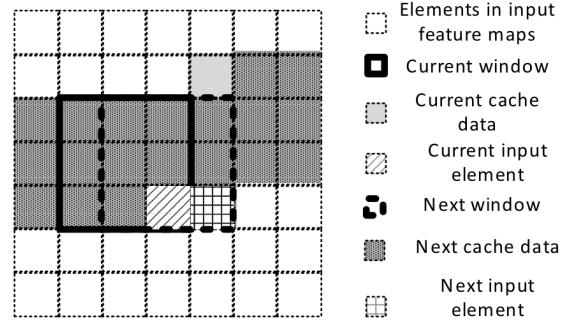

Figure 5 Data reuse in adjacent windows. Totally k=3 lines of input feature maps are cached on chip, which is optimized for data loading. Only one next input element is needed for the new window.

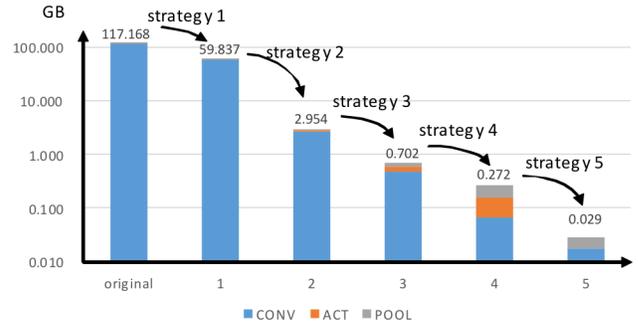

Figure 6 Memory traffic reduction for strategies..

Figure 4 Data flow for strategies 2-4. Solid squares in input feature maps describe current co-located windows, while dash squares stand for next adjacent co-located windows. Reuse in a feature map corresponds black overlapped part. The active windows are reused by co-located filters (green part), which corresponds strategy 2. The data flow of strategy 3 and 4 are shown as red and blue arrows, respectively.

Figure 4 illustrates the data reuse of an active window. $m$ Co-located filters corresponding the current active window are processed (green part) by CUs continuously. The active window is kept with a $32 \times k \times k$-bit register so that CUs can access this window simultaneously.

This data-reuse strategy can reduce input memory traffic to $1/m$ of the conventional. The normalized traffic decreases to $(114.7GB/2/128+2.3GB)/ 28.6GFlop=96.1MB/GFlop$.

### D. Strategy 3: avoid off-chip traffic of intermediate results

The strategy 2 generates intermediate results of output elements, while the strategy 3 accumulates the intermediate to avoid the off-chip traffic. For the $2^{nd}$ convolutional layer of the AlexNet [4], 198.2MB memory space is needed.

Figure 4 includes an efficient data flow based on accumulation of the co-located outputs. Accumulators are designed for $m$ outputs, which correspond all output feature maps. To accumulate outputs, the selection of input windows must jump between co-located windows, since those windows contribute to the same group of output elements. Particularly in our design, we determine the order of jumps from the first to the last. After that, calculations related to one group of co-located output elements are finished. Hence these elements can be streamed out for the next new elements.

The accumulators for the co-located elements take up $32 \times m$-bits registers. Only 1/n original data are streamed out after accumulation. As a result, the normalized traffic decreases to $(114.7GB/2/128+2.3GB/48)/ 28.6GFlop=17.3MB/GFlop$.

### E. Strategy 4: reuse input elements in a feature map

Adjacent $k \times k$-size windows in convolutional layers have 2k overlapped elements. In our consideration, most of overlapped elements can be reused for next window if we promise flow of window are adjacent as far as possible.

Figure 5 shows the overlapped relation between 3×3 windows. we decide the window swift from the left to the right. Due to the dependency, the k lines where the current window is located can be cached on chip. When next window is required, only one extra new element rather than the whole window is streamed in. To avoid conflicts with strategy 2 and 3, we decide to cache k-lines in all input feature maps. As shown in Figure 4, when one group of co-located output elements are finished, we turn next adjacent co-located window of input feature maps.

Cache is required for all input feature maps. Hence a total of $m \times k \times Cin \times 4B$ on-chip memory are needed, where $Cin$ means the number of columns of input feature maps. As the typical convolutional layer shown in Table 1, traffic of inputs can be reduced to $1/(5 \times 5)$ of the original with the memory cost of 14.16KB SRAM.

### F. Strategy 5: super layer integration

Strategies 2-4 reuse data in convolutional layers. Between layers, redundancy of load/save of feature maps still exists, since the output and input feature maps of adjacent layers are exactly same. Moreover, the structure of CNNs can be decomposed into several cascades of convolutional layers, followed with activation and pooling layers. As a result, we combine each cascade of the three layers as a super layer to reduce memory traffic between them.

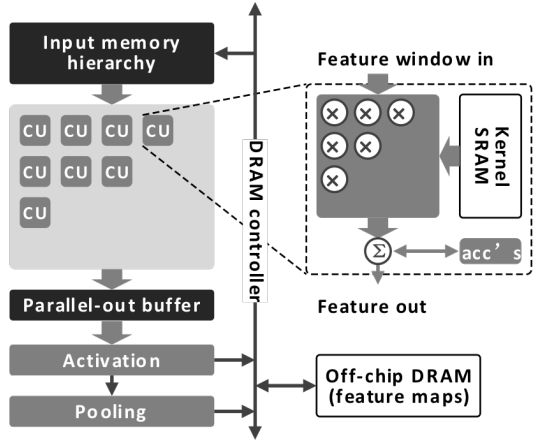

Figure 7 Architecture of CNN-MERP. This architecture works for FP of CNN. The reconfigured architectures for BP and KU contain differences, which are discussed in Section VI.

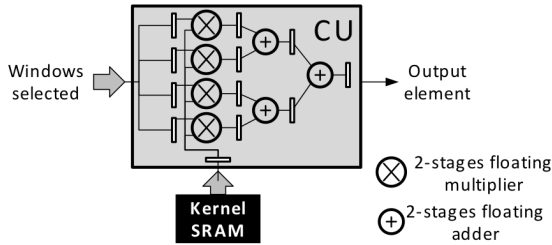

Figure 8 Hardware implementation of computational units for $k \times k$ kernel. Here, $k=2$ in the typical layer of AlexNet, $k=3$.

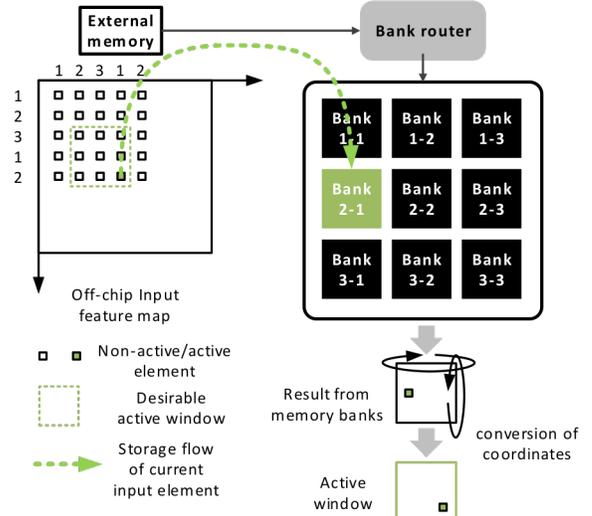

Figure 9 $k \times k$ kernel example for input memory hierarchy. $k=3$. The input element is cached into the memory bank with the same coordinates. Since the location of the window is shifted, while the locations of banks in window are fixed, we use a module to convert the coordinates.

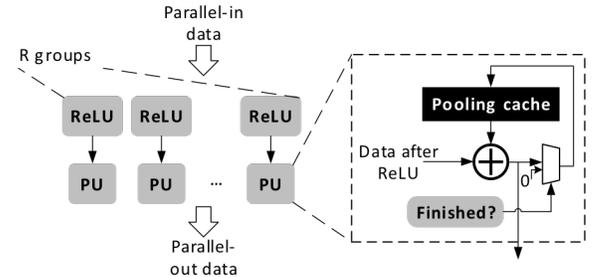

Figure 10 Implementation of pooling engine. $R=2$ in our design.

Figure 6 shows the reduction based on previous strategies. The total memory traffic is reduced by 3976x relative to a no-cache design. The strategies above can promise that all input/output feature maps are loaded/saved only once. The normalized bandwidth requirement with above strategies decreases to $29.6MB/28.6GFlop=1.01MB/G$Flop. This efficiency is enough to support the acceleration of the FPGA resources with a mainstream memory interface.

## V. IMPLEMENTATION OF CNN-MERP

### A. Processor overview

Our implementation incorporates the functionalities of super layers, which are mapped into the FPGA. Figure 7 shows an overview of CNN-MERP. All kernels are loaded into the SRAM prior to the execution of a batch. For the convenience of kernel transfer, each computational unit is equipped with one SRAM to store all related data. We also design an input memory hierarchy corresponding the strategy 2-4 introduced in section IV.

### B. Computational units

The implementation of computational units is based on the parallelism evaluated in section 4. As shown in Figure 8, We arrange $k \times k$ floating-point multipliers in parallel, with $k \times k-1$ floating-point adders followed. In order to get higher throughput, we pipeline adders and multipliers into two stages.

### C. Input memory hierarchy

An imbalanced-data-width problem exists according to strategy 2 and 4. On one hand, in every clock cycle, input memory only gets co-located elements for one location. On the other, CUs need $k \times k$-location input windows. Hence we use $k \times k$ two-port SRAM banks to cache the input data. As shown in Figure 9, in each clock cycle, one of the SRAM banks gets a new element from the router, meanwhile each SRAM bank contributes one element as the input window. As windows shifted, we implement a logic circuit to restore the location of windows.

### D. Activation and pooling

Figure 10 shows the implementation of activation and pooling layers. ReLUs and Pooling units (PUs) are designed for calculation of each element. Because co-located output elements from the convolutional layers are generated simultaneously, we employ $R$ ReLU and PU to calculate co-located elements in feature maps in parallel.

The parallelism of ReLU and PU depends on throughput of CUs. In the typical 2nd layer of AlexNet shown in Table 1, we implement 16 CUs, which takes $48 \times 128/16=384$ clock cycles to work out 128 parallel-out elements. Hence PUs have 384 clock cycles to finish subsampling of 128 output elements. To ensure throughput, we implement $R=2$ ReLUs and PUs.

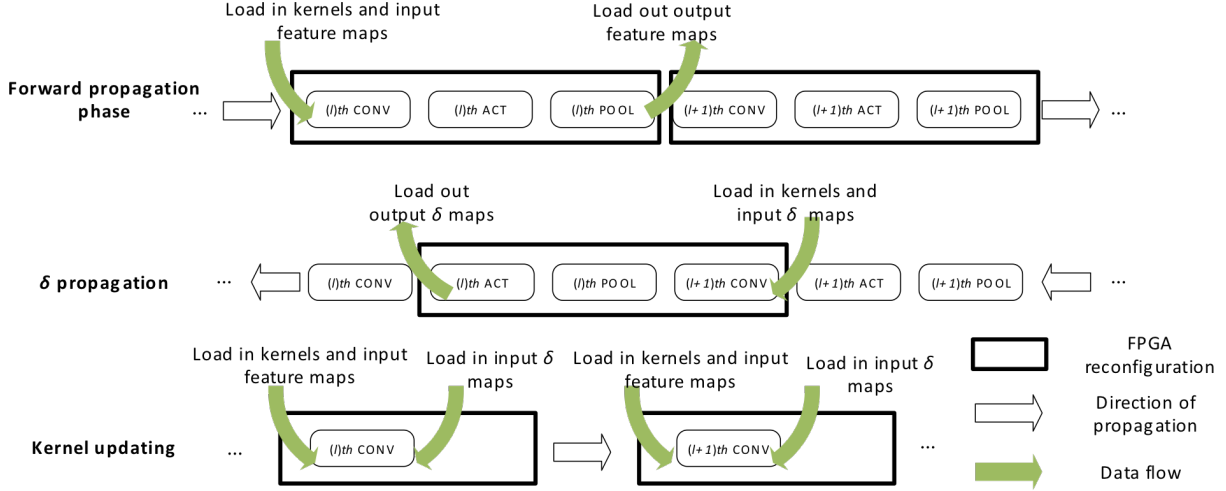

Figure 12 Illustration of data operating flow diagram. Operations in a bold black box are processing on one FPGA hardware system. Green arrows mean off-chip data traffic. In the same layer, feature maps, kernels, $\delta$ maps keep the same in different phases, i.e. the input feature maps of FP are also passed into KU step. The output $\delta$ maps of $\delta$ propagation is the input of KU.

## VI. PROCESSOR RECONFIGURATION FOR FP AND BP

### A. Combining two levels of reconfigurability

The compatibility of different layers for the same hardware is a difficult problem, to which there are two solutions in general. One solution takes advantage of reconfigurable functionality of FPGA, but results in the cost of time on reconfiguration. The other is based on logic circuits to support the new desirable data flow on chip. This solution has an unsatisfied utilization of computational resources when the size of kernels changes. Because the size of CUs is hard to change, we must implement a large-size CU to support all smaller-size kernels, which leaves much idle multipliers and adders when size of kernel is small. Hence we combine two levels of configurability:

When the sizes of kernels do not change, we implement a module to alter the data flow so that the idleness of CUs is omitted as far as possible.

When the sizes of kernels or propagation styles change, we load a new bit stream configuration into the FPGA, so that the new CUs keep the same size of kernel in current super layer.

### B. Logic based reconfiguration

Logic modules are implemented to monitor the current active input and output feature maps, which prevent CUs from invalid inputs/outputs. The invalid is resulted from the mismatching of the real number and the supported number of input/output feature maps. A naive idea is we treat all invalid inputs as 0. This method can get the correct results of computation, however, many zeros are passed to CUs, which reduce the effective performance.

Our solution is to control the data flow so that invalid inputs/outputs are not involved in computation. Registers are designed for current index of co-located windows and output elements. The current index must vary in the valid range, i.e. the real number of input and output feature maps. These two numbers are also from the outside and can be updated for a new layer. As a result, CUs deliver a higher effective throughput in different layers. For example, the $3^{rd}$, $4^{th}$ and $5^{th}$ layers in AlexNet contain the same size of kernel so we use the largest-scale layer to support three layers without FPGA based reconfiguration (256 input feature maps and 384 output feature maps). We implement 48 CUs. The efficiency of the directed idea is 100%, 37.5%, 25%, while in our design, the efficiency keeps 100%, 100%, 88.9%.

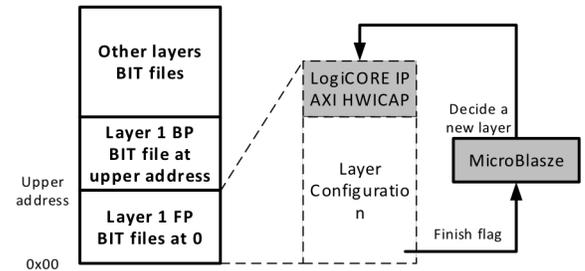

Figure 11 Multiboot reconfiguration flow of FPGA. For our FPGA one version of bit file takes up 11.4MB.

### C. Multi-boot based reconfiguration of FPGA

Multi-boot based reconfiguration takes advantage of the programmability of FPGA itself. Several editions of bit streams can be accessible by the same FPGA. In [20], Xilinx FPGA can be reconfigured from flash on chip, which results in a different functionality.

Multi-boot is realized by the BPI interface on the FPGA board. Before multi-boot, we store all bit streams, each of which corresponds to the functionality of one or more super layers in the flash and record corresponding addresses. When calculation of one super layer is finished, the MicroBlaze soft microprocessor core controls loading of a next layer. Figure 13 shows the flow of reconfiguration. The HWICAP IP core of corresponding layer is triggered by the MicroBlaze soft microprocessor when the next bit stream should be loaded.

Current Xilinx Kintex 325T FPGA takes about 0.7s to process the $2^{nd}$ super layer of the AlexNet with a batch size of 128. With 16-bit BPI interface at 66MHz for transmitting bit files, about 0.087s are spent for receiving data for reconfiguration of one super layer. The overhead of reconfiguration is 11%.

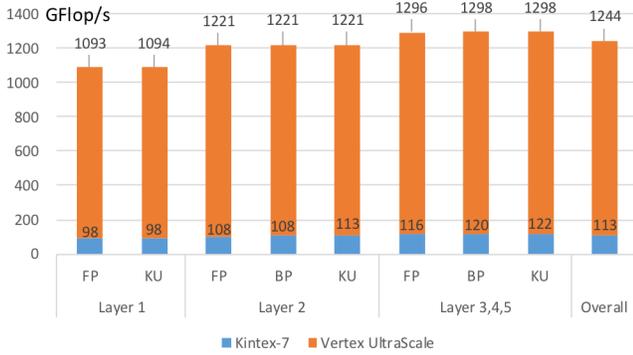

Figure 13 Throughput increment from FPGA extensions. Based on the

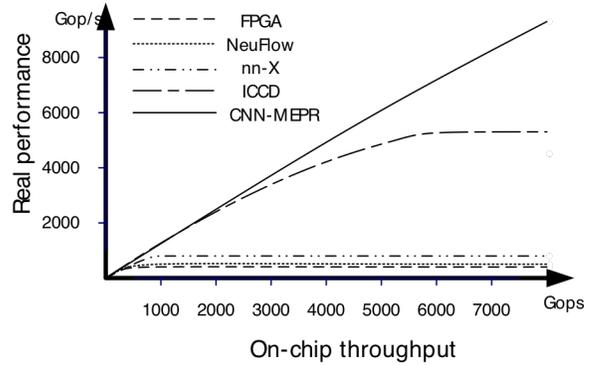

Figure 14 Comparison of real attainable throughput in different works

### D. Reconfiguration flow for FP/BP

The CNN-MERP supports both of FP and BP with the data flow shown in Figure 12. FP comprises the $l^{th}$ super layer which is combined with $l^{th}$ convolutional layer, $l^{th}$ activation layer and $l^{th}$ pooling layer. While BP combines the $(l+1)^{th}$ convolutional layer, $l^{th}$ pooling layer and $l^{th}$ activation layer as the $(l+1)^{th}$ super layer. The different order in BP is due to the inversed layers and the data dependency for the KU. Since the output $\delta$ maps are inputs of kernel updating, super layers should be separated by external traffic of $\delta$ so that $\delta$ can be stored outside. Due to the design above, $\delta$ of the 1$^{st}$ layer is calculated in the 2$^{nd}$ super layer. Hence there is no $\delta$ propagation of the 1$^{st}$ layer.

## VII. IMPLEMENTATION RESULTS

The CNN-MERP as described is implemented on Xilinx Kintex-7 xc7k325tffg900-2 platform. The development was done with Vivado 2015. We program different bit streams for every super layer in AlexNet [4]. Each super layer is operated separately. Since the kernel size of the 3rd 4th and 5th super layers are the same, the largest-scale 3$^{rd}$-super-layer design is compatible to any of them.

The CNN-MERP has well extensibility which makes parallelism can be easily improved on a larger platform. Table 2 shows implementation results of the 2$^{nd}$ layer. We use a larger Xilinx Vertex UltraScale xcvu440 platform to extend our design. Figure 13 shows the comparison of throughput after synthesis in each layer. The throughputs are increased to over 10x. The overall throughput of UltraScale FPGA is 1244GFlop/s, which is 5.48 times larger than state-of-the-art FPGA works. Since the two-level reconfiguration are utilized, the throughput keeps high for all layers.

Table 3 compares the performance of memory traffic reduction. Our implementation and Eyeriss both take AlexNet as an example to optimize memory traffic. The CNN-MERP outperform in each layer because of strategies proposed and the overall normalized bandwidth requirement is 1.94MB/GFop, which is 55.0% lower than Eyeriss. The first layer performs larger requirement because of the stride exists in convolutional layer. The stride means the distances of two adjacent windows. Usually, stride equals to one which is performed in 2$^{nd}$ -5$^{th}$ layer.

Table 3 Comparison with Eyeriss on external bandwidth requirement. Since the training process requires 32-bit operation, Double normalized Bandwidth in this table are needed for Eyeriss-style processor.

| Layer | Operations (Gop/s) | Eyeriss [17] (MB/Gop) Normalized BW of FP (16-bit) | CNN-MERP(32-bit) (MB/GFlop) | | |
|---|---|---|---|---|---|
| | | | Normalized BW of FP | Normalized BW of $\delta$P | Normalized BW of KU |
| 1 | 27.01 | 7.11 | 4.18 | --- | 8.36 |
| 2 | 57.34 | 3.13 | 1.01 | 4.25 | 2.29 |
| 3 | 38.27 | 4.26 | 1.45 | 3.37 | 1.45 |
| 4 | 28.74 | 4.21 | 2.31 | 2.31 | 2.31 |
| 5 | 19.14 | 4.13 | 1.98 | 2.89 | 2.89 |
| Total | 170.50 | 4.31 | 1.94 | 3.45 | 3.92 |

Because the stride of 1$^{st}$ layer takes 4, one output is obtained by 16 input elements. Hence larger requirement is needed to support the same throughput. Our design support both FP and BP. Due the structure of CNNs, more data are needed in $\delta$ propagation and KU. Hence the normalized bandwidth requirement is not as well as that of FP.

Table 2 Implementation results of 2$^{nd}$ super layer in AlexNet

| Resource (slices) | Kintex Kintex-7 | | | Vertex UltraScale | | |
|---|---|---|---|---|---|---|
| | FP | $\delta$P | KU | FP | $\delta$P | KU |
| LUT | 182367 | 178435 | 173195 | 1505983 | 1356150 | 1302193 |
| FF | 121498 | 114082 | 117959 | 854134 | 868097 | 850025 |
| BRAM | 213 | 238 | 209 | 1526 | 1838 | 1498 |
| DSP | 413 | 408 | 405 | 2848 | 2848 | 2848 |

Table 4 Performance comparison with other works.

| Work | FPGA 2015 [9] | NeuFlow [10] | nn-X [11] | Eyeriss [17] | ICCD 2013 [18] | Our implementation |
|---|---|---|---|---|---|---|
| Precision | 32bit float | 16bit fixed | 32bit float | 16bits fixed | --- | 32bits float |
| FPGA / ASIC | Xilinx Vertex 7 VX485T | Xilinx Vertex 6 VLX240T | Xilinx Zynq XC7Z045 | TSMC 65nm | Xilinx Vertex6 ML-605 | Xilinx Kintex 7 K325T/ Vertex xc7v2000 |
| Frequency | 100MHz | 200MHz | 142MHz | 100~250MHz | 150MHz | 137.0MHz/189.0MHz |
| Throughput | 61.62GFlop/s | 160Gop/s | 227Gop/s | 33.6~84.0GFlop/s | 42Gop/s | 113GFlop/s/1244GFlop/s |
| Normalized bandwidth requirement | 25.15MB/GFlop | 24.7MB/Gop | 20MB/GFlop | 4.31MB/Gop | 3.57MB/Gop | 1.94MB/GFlop |

The comparison with other related works is shown in Table 4. Our implementation support both FP and BP, while other works only take FP into consideration. Taking the effect of stride into account, the overall normalized bandwidth requirement of FP is 45.7% lower than previous FPGA implementations. We also evaluate the maximum acceptable throughput in extensions, which is shown in Figure 14. With the support of current DDR4 memory (highest 19.2GB/s), three works are unable to reach 1Top/s. [18] meets the bottleneck at 5.37Top/s. CNN-MERP can reach 9.90Top/s.

## VIII. CONCLUSION

In this paper we focus on memory hierarchy in the CNN-MERP, a CNN processor. Recently for high-throughput hardware solution for CNNs, memory traffic becomes the bottleneck of acceleration. CNN-MERP is not only high-throughput but also memory-efficient. Our implementation reaches the performance of 113GFlop/s and 1.94MB/GFlop. The normalized bandwidth requirement is 45.7% lower than the state-of-the-art work. We also extend the design to a larger scale. As a result, 1244GFlop/s is achieved, which is 5.48 times larger than previous works.

CNN-MERP can process both forward and backward propagation in CNNs. We can fully utilize hardware resources across layers with two-level reconfiguration. Our future work is to enhance the computational density in hardware to attain higher speed of acceleration for CNNs. Another problem is to utilize multiple FPGAs to accelerate CNNs.